\documentclass[10pt,twocolumn,letterpaper]{article}

\usepackage{iccv}
\usepackage{times}
\usepackage{epsfig}
\usepackage{graphicx}
\usepackage{amsmath}
\usepackage{amssymb}

\usepackage{booktabs}
\usepackage{float}
\usepackage{bbding}
\usepackage{pifont}
\usepackage{wasysym}
\usepackage{multirow}
\usepackage{amsfonts}
\usepackage{subfigure}
\usepackage{color}
\usepackage[breaklinks=true,bookmarks=false]{hyperref}

\iccvfinalcopy 

\def\iccvPaperID{8998} 

\ificcvfinal\fi

\begin{document}

\title{Motion-Focused Contrastive Learning of Video Representations\thanks{{\small This work was performed at JD AI Research.}}}

\author{Rui Li $^{\dag}$, Yiheng Zhang $^{\ddag}$, Zhaofan Qiu $^{\ddag}$, Ting Yao $^{\ddag}$, Dong Liu $^{\dag}$, and Tao Mei $^{\ddag}$ \\
         $^{\dag}$ University of Science and Technology of China, Hefei, China~~~
         $^{\ddag}$ JD AI Research, Beijing, China\\
{\tt\small liruid@mail.ustc.edu.cn, \{yihengzhang.chn, zhaofanqiu, tingyao.ustc\}@gmail.com}\\
{\tt\small dongeliu@ustc.edu.cn, tmei@live.com}
}

\maketitle
\ificcvfinal\thispagestyle{empty}\fi

\begin{abstract}
   Motion, as the most distinct phenomenon in a video to involve the changes over time, has been unique and critical to the development of video representation learning. In this paper, we ask the question: how important is the motion particularly for self-supervised video representation learning. To this end, we compose a duet of exploiting the motion for data augmentation and feature learning in the regime of contrastive learning. Specifically, we present a Motion-focused Contrastive Learning (MCL) method that regards such duet as the foundation. On one hand, MCL capitalizes on optical flow of each frame in a video to temporally and spatially sample the tubelets (i.e., sequences of associated frame patches across time) as data augmentations. On the other hand, MCL further aligns gradient maps of the convolutional layers to optical flow maps from spatial, temporal and spatio-temporal perspectives, in order to ground motion information in feature learning. Extensive experiments conducted on R(2+1)D backbone demonstrate the effectiveness of our MCL. On UCF101, the linear classifier trained on the representations learnt by MCL achieves 81.91\% top-1 accuracy, outperforming ImageNet supervised pre-training by 6.78\%. On Kinetics-400, MCL achieves 66.62\% top-1 accuracy under the linear protocol. Code is available at https://github.com/YihengZhang-CV/MCL-Motion-Focused-Contrastive-Learning.
\end{abstract}

\vspace{-0.2in}
\section{Introduction}
\vspace{-0.1in}
The key difference between video and image is the dimension of time, which derives a particular form of motion information in a video. The state-of-the-art works often delve into motion in different ways, e.g., long/short term dependencies~\cite{qiu2019learning,wang2018non,li2019long}, temporal structure/orders \cite{lee2017unsupervised,li2021representing,xu2019self}, and temporal pooling~\cite{wang2016temporal,yang2020temporal}, to enhance video understanding. The underlying foundation behind these advances generally originates from the improvement of representation learning via the exploration of motion information. Most recently, self-supervised representation learning is gaining significant momentum~\cite{cai2020joint,chen2020simple,he2020momentum}, and the number of self-supervised learning papers practically exploded. In particular, contrastive learning, as a memory-based self-supervised learning approach, is extended to video domain \cite{han2020self,yao2020seco}  and further closes the gap between self-supervised and supervised video representation learning. A valid question then emerges as how important is the motion for self-supervised video representation learning?

In an effort to answer the question, we look into the problem, in the context of contrastive learning, from two different perspectives: 1) leveraging motion information in achieving data augmentations, and 2) taking motion into account in the optimization of feature learning. In a video, the motion of different regions is inherently various and the velocity of motion measures the rate of change in position of the region with respect to a frame of reference. In general, the regions with larger velocities have much richer information and are potentially more advantageous for contrastive learning. As a result, we capitalize on motion information from both spatial and temporal dimensions to carefully sample the sequences of patches across frames, i.e., tubelets, as augmentations, and examine how the first issue affects self-supervised video representation learning. To study the second one, we encourage the grounding of motion information explicitly in feature learning by aligning gradient maps of the convolutional layers to motion (optical flow) maps. As such, feature learning executes the optimization with respect to motion information.

To consolidate the idea of exploring the motion information in video sequence for self-supervised video representation learning, we present a novel Motion-focused Contrastive Learning (MCL) method. Specifically, we leverage unsupervised TV-L1 algorithm \cite{zach2007duality} to extract the dense optical flow of each frame in a video and compute the motion boundaries as in \cite{dalal2006human} to obtain the motion map. A video is divided into a set of fixed-length video clips and the spatio-temporal motion map (ST-motion) of each video clip consists of the sequential motion maps of all the frames in the clip. MCL then performs a motion-focused spatio-temporal sampling to select tubelets as data augmentations. Technically, MCL applies a 3D average pooling on the spatio-temporal motion map to measure the clip-level motion, which indicates the degree of motion of each clip. The clips with relatively large clip-level motion are chosen as the clip candidates for temporal augmentation. Next, MCL employs a temporal pooling on the spatio-temporal motion map of each clip candidate to estimate the motion from spatial viewpoint (S-motion) and localize the spatial patches, which are temporally consistent across frames, as tubelets. Furthermore, in feature learning, MCL extracts the spatio-temporal motion map of each tubelet and executes a spatial/temporal pooling on such spatio-temporal motion map to output the motion map from temporal/spatial viewpoint (T/S-motion) of the tubelet. The gradient with regard to the feature map of a convolutional layer by back-propagation is produced in spatial, temporal and spatio-temporal manner, respectively, to align with S-motion, T-motion, ST-motion through minimizing the mean squared error in between. MCL integrates the alignments into contrastive learning framework as constraints in addition to InfoNCE loss.

The main contribution of the work is the proposal of leveraging motion information to boost self-supervised video representation learning on the recipe of contrastive learning. This leads to the elegant views of how to effectively sample spatio-temporal augmentations in terms of motion, and how to integrate motion information into the optimization of feature learning, which are problems not yet fully understood. We demonstrate that our self-supervised method MCL surpasses ImageNet supervised pre-training on two video benchmarks and the experiments on two downstream video tasks also validate our MCL.

\section{Related work}
\textbf{Unsupervised video representation learning} aims to explore the intrinsic properties in unlabeled videos to learn video representation.
The research in this direction has proceeded along two different dimensions: transformation-based methods \cite{benaim2020speednet,jing2018self,kim2019self}, and temporal context-based approaches \cite{diba2019dynamonet,han2019video,han2020memory,lee2017unsupervised,xu2019self}.
Transformation-based methods are optimized to predict the transformation parameters from the transformed videos. Jing \etal \cite{jing2018self} introduce a pretext task to estimate the rotation angle applied to videos. 3D ST-puzzle \cite{kim2019self} proposes a self-supervised task to classify the arrangement of cropped spatio-temporal pieces. SpeedNet \cite{benaim2020speednet} learns video representation by estimating the speed or pace of the transformed videos. 
Temporal context-based approaches focus on exploring the natural temporal relation as supervision.
In~\cite{lee2017unsupervised,misra2016shuffle,xu2019self}, predicting the order of frames or video clips drives the learning of spatio-temporal representation.
The video representation in \cite{han2019video,han2020memory} are learnt by a dense encoding of spatio-temporal blocks to recurrently generate the future representations. Dynamonet \cite{diba2019dynamonet} directly takes the reconstruction of future frames as the pretext task.

\textbf{Learning with motion.} Motion information, as the representation of the changes over time, has been studied by researchers for a long time.
For instance, in \cite{liu2017robust,tsai2016video,zhu2017flow}, the optical flow is utilized to propagate the adjacent frame representations.
In \cite{simonyan2014two}, the famous two-stream architecture is devised by applying two 2D CNN architectures separately to RGB frames and optical flows for action recognition.
The idea of two-steam architecture is also explored from the perspective of knowledge distillation~\cite{crasto2019mars,stroud2020d3d, xiao2021space}.
Moreover, Wang \etal \cite{wang2019self} devise a self-supervised pretext task by estimating the motion in unlabeled videos.

\textbf{Contrastive learning} recently has received intensive research attention due to its promising result on self-supervised visual representation learning. The contrastive loss is devised to return low values for similar pairs and high values for dissimilar pairs, which encourages invariant features on the low dimensional manifold.
In an early work \cite{wu2018unsupervised}, the constrastive learning is formulated as instance-level classification, and the previously computed features are stored in a memory bank to acquire more negative samples.
Momentum Contrast (MoCo) \cite{he2020momentum} builds a dynamic memory bank to maintain a large number of negative samples with a moving-averaged encoder.
\cite{chen2020simple,tian2020makes} further investigate the importance of data augmentations and non-linear in contrastive learning.

In short, our work in this paper mainly focuses on improving constrastive learning for video representation learning through the involvement of motion information. The most closely related works are the contrastive learning frameworks for video representation learning \cite{han2020self,qian2020spatiotemporal,wang2020self,yao2020seco}.
SeCo \cite{yao2020seco} and Pace Prediction \cite{wang2020self} combines contrastive loss with order prediction and pace estimation pretext task, respectively.
CoCLR \cite{han2020self} exploits the complementary information from optical flow and introduces a co-training scheme to improve the spatio-temporal representation. CVRL \cite{qian2020spatiotemporal} studies what makes the good data augmentation for video self-supervised learning. Our method differs from these works in that we exploit the motion information in contrastive learning framework from the perspectives of both data augmentation and representation learning.

\begin{figure*}[!tb]
       \centering {\includegraphics[width=0.875\textwidth]{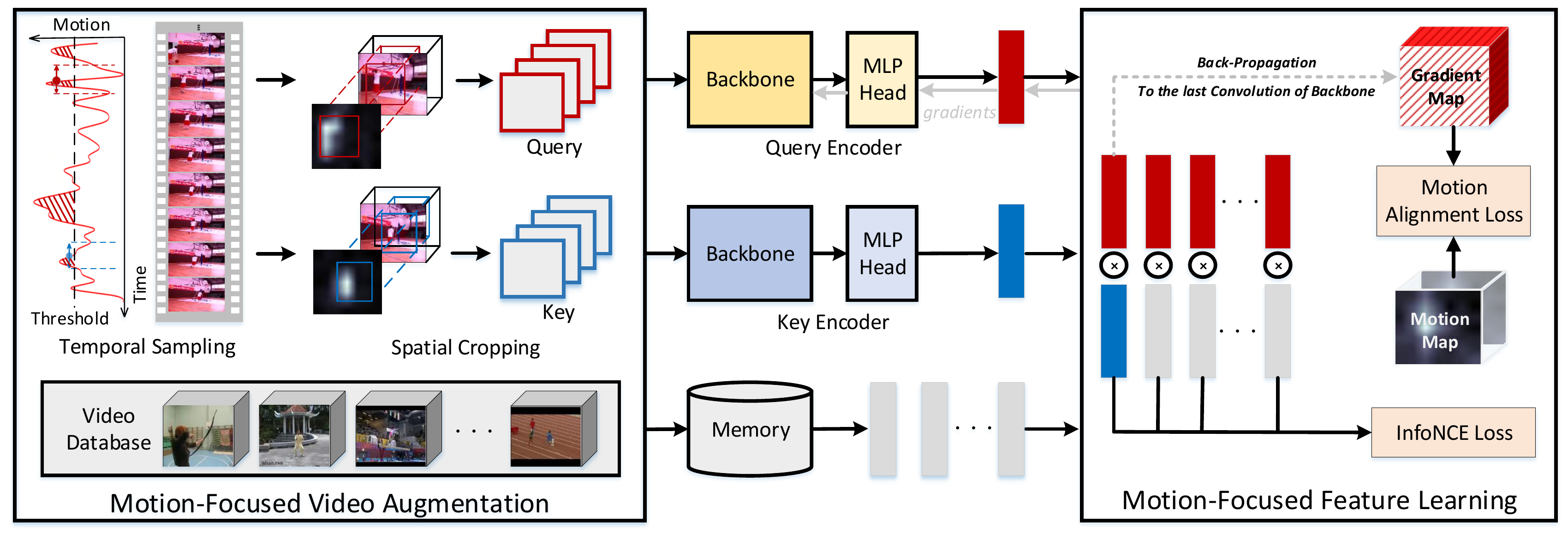}}
       \vspace{-0.05in}
       \caption{An overview of our \textbf{Motion-focused Contrastive Learning (MCL)} framework. MCL facilitates the basic contrastive learning framework by the two novel designs: motion-focused video augmentation and motion-focused feature learning. Specifically, the \textbf{motion-focused video augmentation} exploits the pre-estimated motion map to produce the 3D tubelets with rich motion information as the inputs to backbone network. The augmentation process can be divided into two parts, i.e., temporal sampling to filter out the relatively static clips, and spatial cropping to select the spatial regions with high moving velocity. For \textbf{motion-focused feature learning}, a new motion alignment loss is devised to enforce the backbone network to focus more on the positions with higher temporal dynamics by aligning the magnitude of gradient and motion map in each position. The motion alignment loss is integrated into the contrastive learning framework as constraints in addition to InfoNCE loss. The whole MCL framework is jointly optimized in an end-to-end manner.}
       \label{framework}
       \vspace{-0.2in}
\end{figure*}

\section{Motion-Focused Contrastive Learning}
The basic idea of Motion-focused Contrastive Learning (MCL) is to facilitate self-supervised video representation learning by the awareness of motion information. MCL first pre-estimates the motion map to measure the pixel-level movement in video data. The regions with larger velocities generally contain richer information (e.g., moving object, action, interaction), and therefore have higher priority in MCL. Specifically, our framework integrates the motion map into contrastive learning from the perspectives of both data augmentation and feature optimization. Figure~\ref{framework} illustrates the overview of our MCL.

\subsection{Motion Estimation}
\begin{figure}[!tb]
       \centering
       \subfigure[\scriptsize Input Video]{
              \label{fig:motion_map:a}
              \includegraphics[width=0.12\textwidth]{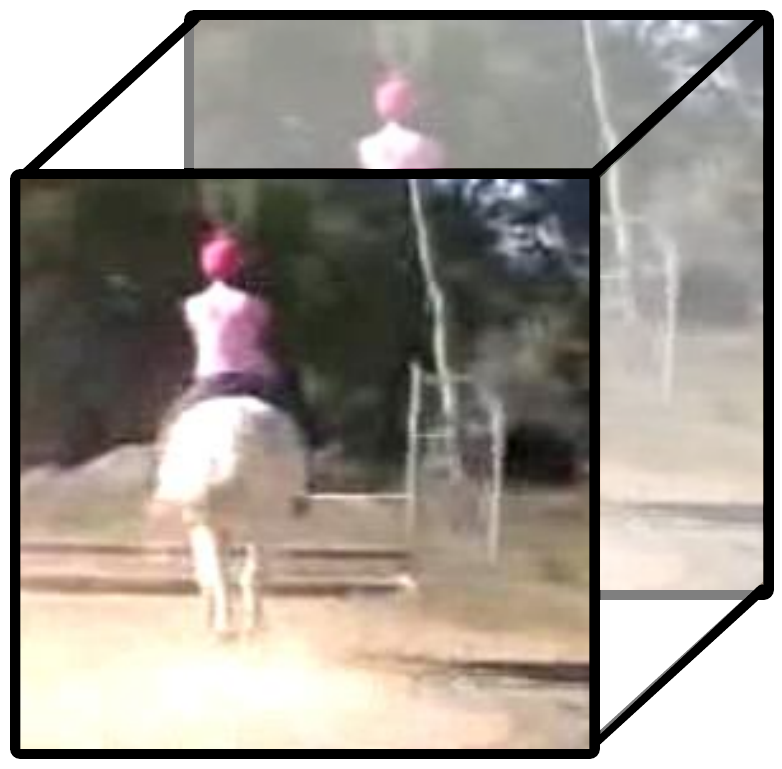}}
       \hspace{0.1in}
       \subfigure[\scriptsize Optical Flow]{
              \label{fig:motion_map:b}
              \includegraphics[width=0.12\textwidth]{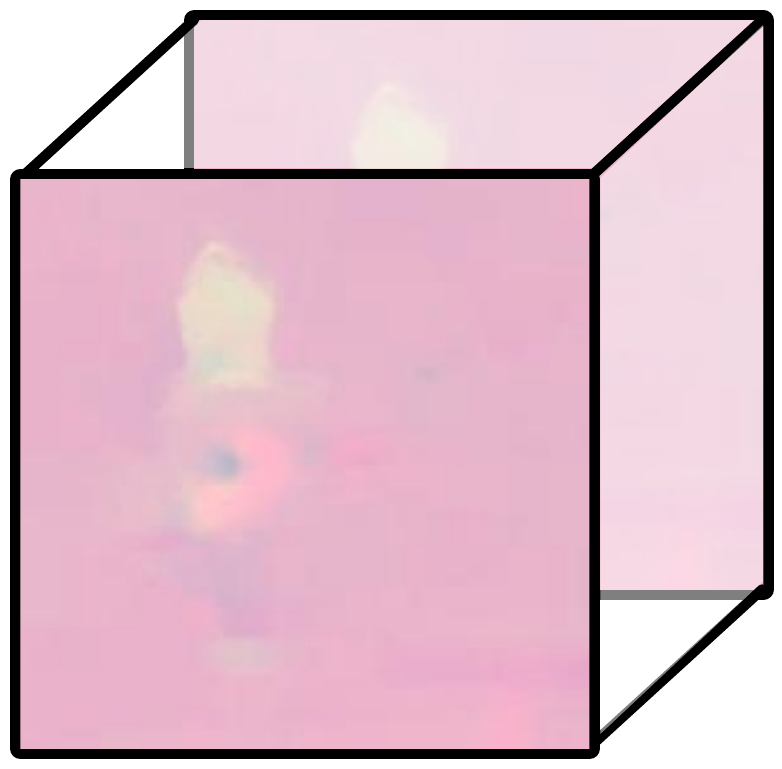}}
       \hspace{0.1in}
       \subfigure[\scriptsize Motion Boundary]{
              \label{fig:motion_map:c}
              \includegraphics[width=0.12\textwidth]{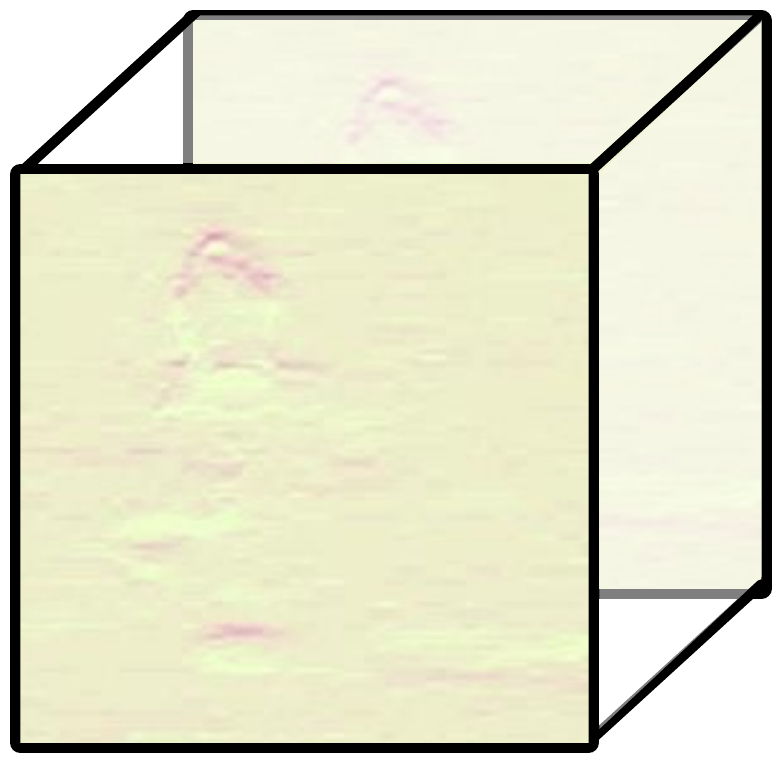}}
       \hspace{0.1in}
       \subfigure[\scriptsize Motion Map]{
              \label{fig:motion_map:d}
              \includegraphics[width=0.12\textwidth]{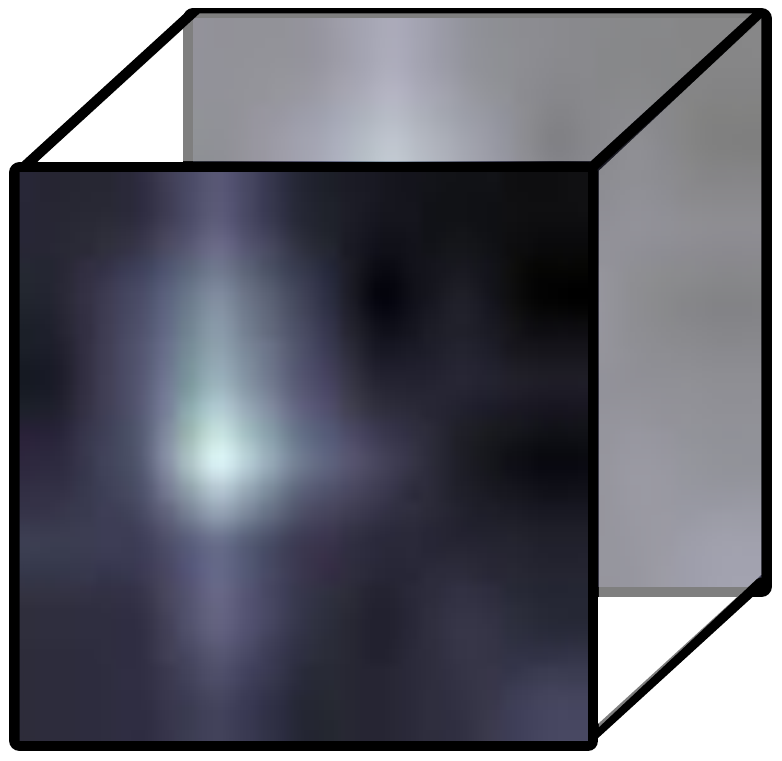}}
       \hspace{0.1in}
       \subfigure[\scriptsize Valuable MI]{
              \label{fig:motion_map:e}
              \includegraphics[width=0.12\textwidth]{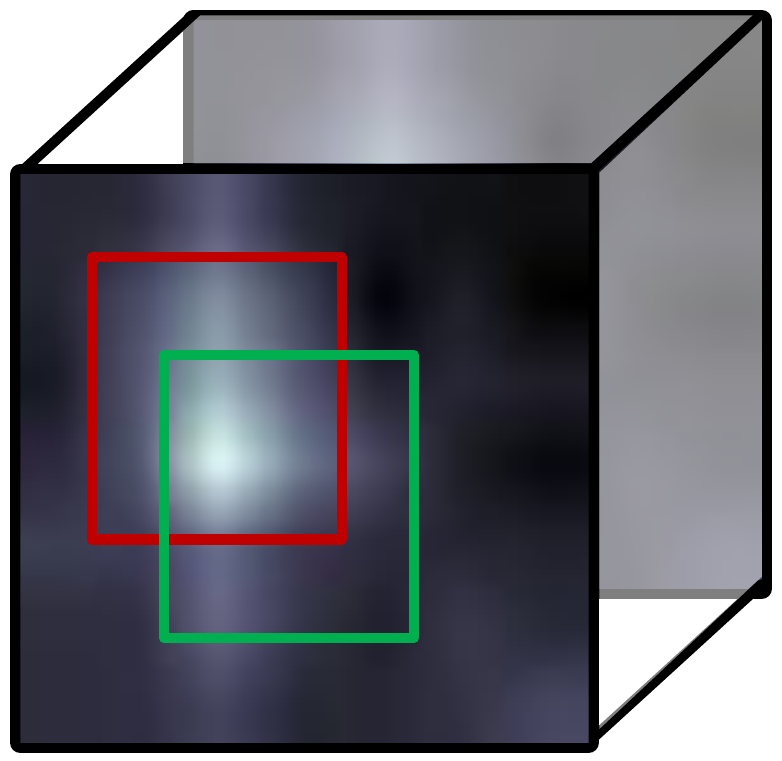}}
       \hspace{0.1in}
       \subfigure[\scriptsize Nuisance MI]{
              \label{fig:motion_map:f}
              \includegraphics[width=0.12\textwidth]{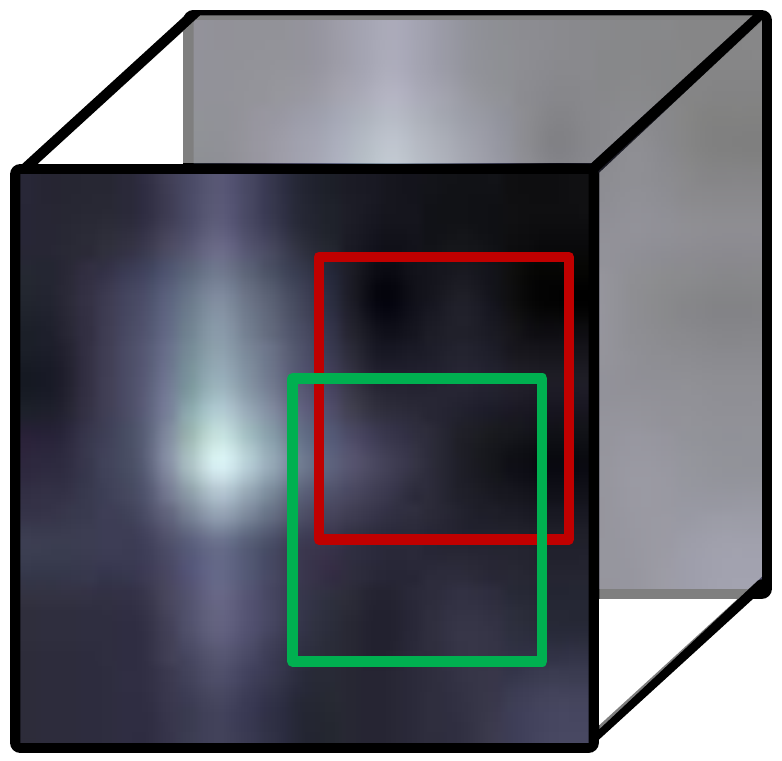}}
       \caption{\small The example of (a) input video sequence, (b) optical flow, (c) motion boundary, (d) motion map, (e) valuable mutual information in the region with large movement and (f) nuisance mutual information in the relatively static region. The red box and green box represent two views generated from the identical instance for contrastive learning.}
       \label{fig:motion_map}
       \vspace{-0.20in}
\end{figure}

In order to explore the motion information in video sequence, MCL starts by estimating the magnitude of motion in each region. Given a video of $N$ frames with the resolution of $H\times W$, we first extract the \textbf{optical flow} from each pair of consecutive frames as $\left\{(\boldsymbol{u}_{1},\boldsymbol{v}_{1}), (\boldsymbol{u}_{2},\boldsymbol{v}_{2}), \ldots, (\boldsymbol{u}_{N},\boldsymbol{v}_{N})\right\}$ by TV-L1 \cite{zach2007duality} algorithm. The flow maps $\boldsymbol{u}_{i}, \boldsymbol{v}_{i} \in \mathbb{R}^{H \times W}$ are the horizontal and vertical displacements of each pixel between $i$-th frame and $(i+1)$-th frame. For the last frame, we manually set $(\boldsymbol{u}_{N}, \boldsymbol{v}_{N})=(\boldsymbol{u}_{N-1}, \boldsymbol{v}_{N-1})$. These optical flow maps have been proven effective in capturing temporal dynamics and widely utilized in video classification methods \cite{carreira2017quo,feichtenhofer2016convolutional,qiu2021optimization,yue2015beyond}. Nevertheless, in our case of measuring the magnitude of movement, the results by optical flow may suffer from stability problem due to camera motion. For example, with large camera motion, the static objects or background pixels also show high moving velocity in optical flow. Hence, we calculate the \textbf{motion boundary} proposed in \cite{dalal2006human} as $(\frac {\partial \boldsymbol{u}_{i}} {\partial x}, \frac {\partial \boldsymbol{u}_{i}} {\partial y}, \frac {\partial \boldsymbol{v}_{i}} {\partial x}, \frac {\partial \boldsymbol{v}_{i}} {\partial y})$, i.e., the x- and y- derivatives of optical flow, to eliminate the effect of camera motion. Finally, we define the \textbf{motion map} by accumulating the amplitudes in four motion boundary maps as
\begin{equation}\label{eq:motion_map}
       \small
       \begin{aligned}
              \boldsymbol{m}_{i}=\sqrt{\left(\frac {\partial \boldsymbol{u}_i} {\partial x}\right) ^{2} + \left(\frac {\partial \boldsymbol{u}_i} {\partial y}\right) ^{2} + \left(\frac {\partial \boldsymbol{v}_i} {\partial x}\right) ^{2} + \left(\frac {\partial \boldsymbol{v}_i} {\partial y}\right) ^{2}}
       \end{aligned}~~,
\end{equation}
where $\boldsymbol{m}_{i} \in \mathbb{R}^{H \times W}$ measures only the moving velocity in $i$-th frame and ignores moving orientation. Figure~\ref{fig:motion_map} showcases the input video and the visualizations of optical flow, motion boundary, and motion map. As illustrated in the figure, the motion map is not influenced by camera motion and shows high responses on the de facto moving objects.

To describe the utilization of motion map more clearly, we pre-define three different types of motion map, i.e., ST-motion, S-motion and T-motion, to measure the moving velocity from different aspects. \textbf{ST-motion} stacks the motion map of all frames to produce a 3D volume $\boldsymbol{m}^{\text{ST}} \in \mathbb{R}^{N \times H \times W}$. \textbf{S-motion} and \textbf{T-motion} averagely pool the motion map through temporal dimension and spatial dimension, respectively:
\begin{equation}\label{eq:motion_map}
       \small
       \begin{aligned}
              \boldsymbol{m}^{\text{S}}&=P_{t}(\boldsymbol{m}^{\text{ST}}) ~~ \in \mathbb{R}^{H \times W}~~, \\
              \boldsymbol{m}^{\text{T}}&=P_{s}(\boldsymbol{m}^{\text{ST}}) ~~ \in \mathbb{R}^{N}~~,
       \end{aligned}
\end{equation}
where $P_{t}(\cdot)$ and $P_{s}(\cdot)$ are the pooling operations. Please note that the motion estimation process does not require manual labeling. Hence, these motion maps can be treated as an additional label-free supervision for video data.

\subsection{Motion-Focused Video Augmentation}
\label{sec:MVA}
Then, the video augmentation in MCL is guided by the acquired motion map and focuses more on the regions with large movements. The rationale behind is to select the better views for contrastive learning that could boost the generalization capability of the learnt representations. Specifically, in contrastive learning, a single instance is split into two views to learn an embedding where the views are relatively closer than the views from other instances. The self-supervised learning therefore will benefit from the mutual information (MI) between views. As studied in \cite{tian2020makes}, to improve the generalization ability to downstream tasks, the ``good'' views should contain as much task-relevant information while discarding as much irrelevant information in the input as possible. Unlike the framework in \cite{tian2020makes} that requires the prior knowledge of downstream tasks to select views, MCL exploits the label-free motion map to seek the regions with large movements, which are generally essential for video-related tasks. For example, the views generated in the region with large motion and relatively static region are shown in Figure \ref{fig:motion_map:e} and Figure \ref{fig:motion_map:f}, respectively. The views with large movements are more likely to contain the valuable mutual information such as moving object or action, while the views in relatively static region only contain the information in background pixels that are usually considered as nuisance information in contrastive learning.

Here, we introduce a simple way to ensure the framework focusing more on the motion information by only selecting the 3D tubelets with large movement as preprocessing. To achieve this, \textbf{motion-focused video augmentation} screens the input video volume by two steps of temporal sampling and spatial cropping, meanwhile maintaining the randomness in the traditional augmentation methods.

\textbf{Temporal sampling} selects the clips (i.e., short frame sequences) with large movements in the video.
Taking the T-motion $\boldsymbol{m}^{\text{T}}$ in Eq.(\ref{eq:motion_map}) as the frame-level motion map, the movement of each clip is measured by averaging the frame-level movement in $\boldsymbol{m}^{\text{T}}$. We take the median across all the candidate clips in an identical video as a threshold, and randomly select one clip with magnitude above the threshold.

\textbf{Spatial cropping} further localizes the cropped spatial region in the sampled clip. It first calculates the S-motion $\boldsymbol{m}^{\text{S}}$ within the clip. 
We then take the 90th-percentile in the motion map as a threshold. 
One bounding box that covers the 80\% pixels with higher value than the threshold is randomly selected.
Once the region is determined, we crop the same region of each frame in the clip, which is the same as the temporally consistent spatial augmentation in \cite{qian2020spatiotemporal}.

Please note that this two-step augmentation can also be equivalently implemented by directly seeking the spatio-temporal region with large movement in the ST-motion. Here we separate it into two steps due to the higher efficiency since temporal sampling only selects a small number of candidate tubelets in the first step. Moreover, the two thresholds in temporal sampling and spatial cropping are relative values within each video/clip. We do not compare the movement magnitude across different videos to avoid over-fitting to the videos with large motion.

Based on the produced clips by temporal sampling and spatial cropping, we follow~\cite{qian2020spatiotemporal,yao2020seco} and also employ color-jittering, random scales, grayscale, blur and mirror.

\subsection{Motion-Focused Feature Learning}
In addition to video augmentation, we also take the motion map as an additional supervision to enhance the learnt representation. Particularly, we begin by briefly reviewing instance discrimination objective in contrastive learning. Given an encoded query $\boldsymbol{q}\in\mathbb{R}^d$ and a group of encoded key vectors $\mathcal{K}=\left\{\boldsymbol{k}^{+}, \boldsymbol{k}_{1}^{-}, \boldsymbol{k}_{2}^{-}, \ldots, \boldsymbol{k}_{K}^{-}\right\}$ consisting of one positive key $\boldsymbol{k}^{+}\in\mathbb{R}^d$ and $K$ negative keys $\mathcal{K^{-}}=\left\{\boldsymbol{k}_{j}^{-}\right\}$, where $d$ denotes the embedding dimension. The query and its positive key are usually two views generated from the same instance, while the negative keys are from the other instances. The objective of instance discrimination is to guide the query $\boldsymbol{q}$ to be similar with the positive key $\boldsymbol{k}^{+}$ while it remains distinct to all negative keys $\mathcal{K^{-}}$. Therefore, a prevailing form of contrastive loss is presented in InfoNCE \cite{oord2018representation} based on a softmax formulation:
\vspace{-0.05in}
\begin{equation}\label{eq:nce}
       \small
       \begin{aligned}
              \mathcal{L}_{\text{NCE}}=-\log \frac{\exp \left(\boldsymbol{q}^{T} \boldsymbol{k}^{+} / \tau\right)}{\exp \left(\boldsymbol{q}^{T} \boldsymbol{k}^{+} / \tau\right)+\sum_{i=1}^{K} \exp \left(\boldsymbol{q}^{T} \boldsymbol{k}_{i}^{-} / \tau\right)}
       \end{aligned}~~,
\end{equation}
where the similarity is measured via dot product, and $\tau$ is the temperature hyper-parameter. Here, we follow the recent variant MoCo \cite{he2020momentum} that proposes the use of a queue to track negative samples from past mini-batches. In this way, all the queuing samples from multiple mini-batches serve as negative keys, and the size of mini-batch can be reduced.

The NCE loss in Eq.(\ref{eq:nce}) performs on the encoded tubelet-level representation, which treats each position in the tubelet equally. As discussed in Section \ref{sec:MVA}, the background positions without movements may lead to nuisance information between views. Consequently, we propose a novel \textbf{motion alignment loss} (MAL) to explicitly align the feature/gradient maps of convolutional layer and the magnitude in low-level motion map as an additional supervision. 
Such supervision encourages the network to focus on the positions with large movements in videos.
To achieve this, we devise the following variants of motion alignment loss:

\vspace{0.07in}\noindent (i) \emph{Align feature map:} The simplest way is to align the magnitude of convolutional feature with the motion map. The region with large movement is expected to have high response in the feature map. Formally, let $\boldsymbol{h}_{c}$ denote the $c$-th channel of output feature map from the last convolutional layer of the backbone. The loss function is formulated as the L2 distance between the summation of $\boldsymbol{h}_{c}$ across all channels and ST-motion:
\begin{equation}\label{eq:mal-1}
       \small
       \begin{aligned}
              \mathcal{L}_{\text{MAL-v1}}=\| \langle \sum_{c}\boldsymbol{h}_{c} \rangle - \langle \boldsymbol{m}^{\text{ST}} \rangle \|^{2}_{2}
       \end{aligned}~~,
\end{equation}
where $\langle \cdot \rangle$ is the L2 normalization of the feature/motion map.

\vspace{0.07in}\noindent (ii) \emph{Align weighted feature map:}
Inspired by GradCAM \cite{selvaraju2017grad}, the magnitude of gradient could measure the contribution of each position better. Hence, we calculate the gradient of the similarity between query and its positive key with regard to the convolutional feature as $\boldsymbol{g}_{c}=\frac{\partial \boldsymbol{q}^T \boldsymbol{k}^{+}}{\partial \boldsymbol{h}_{c}}$. Then, the mean value $w_c$ of the gradient map $\boldsymbol{g}_{c}$ can be utilized as the weight of each channel as
\begin{equation}\label{eq:mal-2}
       \small
       \begin{aligned}
              \mathcal{L}_{\text{MAL-v2}}=\| \langle\text{ReLU}(\sum_{c} w_c \boldsymbol{h}_{c})\rangle - \langle\boldsymbol{m}^{\text{ST}}\rangle \|^{2}_{2}
       \end{aligned}~~,
\end{equation}
where the ReLU operation is added to filter out the regions with negative contribution as in \cite{selvaraju2017grad}.

\vspace{0.07in}\noindent (iii) \emph{Align weighted gradient map:} We further consider to align the gradient map with motion map. As such, motion information can directly guide the update of representations. Specifically, we replace the weighted feature map in Eq.(\ref{eq:mal-2}) with weighted gradient map:
\begin{equation}\label{eq:mal-3}
       \small
       \begin{aligned}
              \mathcal{L}_{\text{MAL-v3}}=\| \langle\text{ReLU}(\sum_{c} w_c \boldsymbol{g}_{c})\rangle - \langle\boldsymbol{m}^{\text{ST}}\rangle \|^{2}_{2}
       \end{aligned}~~.
\end{equation}
The comparisons between different loss functions will be discussed in the experiments, and the alignment of weighted gradient map is used as the default motion alignment loss.

\begin{figure}[tp]
       \centering
       \includegraphics[width=0.43\textwidth]{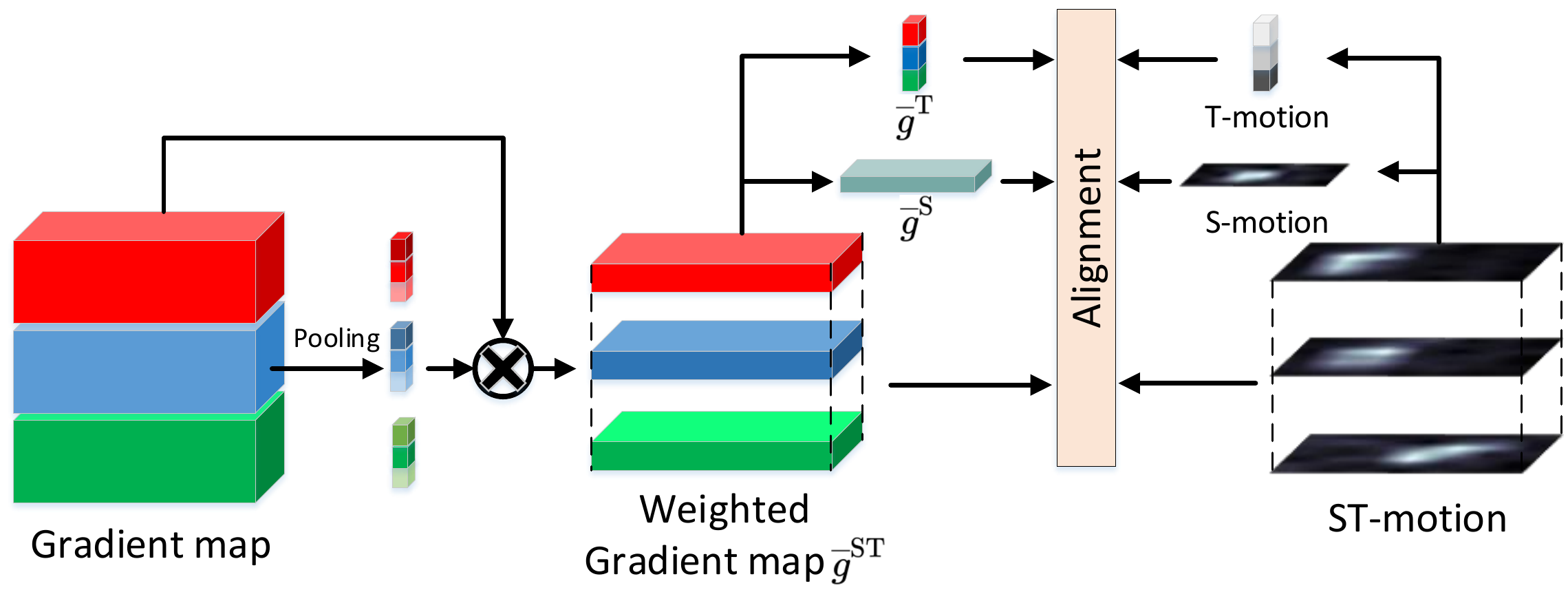}
       \caption{\small An illustration of \textbf{motion alignment loss} which aligns the gradient maps with ST-motion, S-motion and T-motion.}
       \label{gradient_loss}
       \vspace{-0.20in}
\end{figure}

To emphasize the alignment from temporal/spatial aspects, we additionally minimize the distance between spatially/temporally pooled weighted gradient map and T/S-motion, as shown in Figure \ref{gradient_loss}. Specifically, we define the spatio-temporal weighted gradient map as $\overline{\boldsymbol{g}}^{\text{ST}}=\langle\text{ReLU}(\sum_{c} w_c \boldsymbol{g}_{c})\rangle$, and further pool such map through temporal/spatial dimension to produce spatial/temporal weighted gradient map $\overline{\boldsymbol{g}}^{\text{S}}$/$\overline{\boldsymbol{g}}^{\text{T}}$, respectively. The three gradient maps are then aligned to the corresponding motion map:
\begin{equation}\label{eq:mal}
       \small
       \begin{aligned}
              \mathcal{L}_{\text{MAL}}=\|\overline{\boldsymbol{g}}^{\text{S}} - \langle\boldsymbol{m}^{\text{S}}\rangle\|_{2}^{2}+\|\overline{\boldsymbol{g}}^{\text{T}} - \langle\boldsymbol{m}^{\text{T}}\rangle\|_{2}^{2}+\|\overline{\boldsymbol{g}}^{\text{ST}} - \langle\boldsymbol{m}^{\text{ST}}\rangle\|_{2}^{2}
       \end{aligned}~~.
\end{equation}

The overall training objective in our MCL is formulated as a multitask loss by integrating the instance discrimination loss and motion alignment loss, which is written as $\mathcal{L} = \mathcal{L}_{NCE} + \mathcal{L}_{MAL}$, where we empirically treat each loss equally and simply summate the two losses.

\section{Experiments}
We verify the merit of our MCL via various empirical evidences. The evaluation protocols include: 1) linear classification on action recognition, directly trained on frozen pre-learnt features, and 2) fine-tuning the networks initialized from pre-learnt network for downstream tasks of action recognition and video retrieval.

\subsection{Datasets}
UCF101 \cite{soomro2012ucf101} contains about 13K videos from 101 action categories, which are grouped into 9.5K and 3.7K videos for training and testing. HMDB51 \cite{kuehne2011hmdb} consists of 7K videos from 51 action classes, and each split includes 3.5K and 1.5K videos for training and testing. Kinetics-400 dataset \cite{carreira2017quo} is a large-scale action recognition benchmark and contains around 300K videos from 400 action categories. The 300K videos are divided into 240K, 20K and 40K in training, validation and test sets, respectively. Note that the labels of test set are not publicly available and here we report the performances on validation set.

\subsection{Experimental Settings}
\textbf{Implementation Details.}
We exploit the backbones of R(2+1)D based on ResNet-50 \cite{qiu2017learning,tran2018closer} or S3D \cite{xie2018rethinking} plus an MLP head as video encoder for contrastive training. Note that MLP head only works for training and is disabled on downstream tasks. The inputs of tubelets to the backbone are with the size of $16\times 224\times 224$, and each tubelet consists of 16 frames with a temporal stride 2. The MLP head takes 3D global pooling features as the inputs and embeds the feature into 128$d$ via two fully-connected layers ($2048\times 2048$ and $2048\times 128$). We apply L2 normalization to the output vector from the MLP head, which is then used as the encoded feature of query or keys. In the implementations, we set the size of mini-batch and memory to 64 and 131,072, respectively. The momentum coefficient $\alpha$ is fixed to 0.999 for momentum update of video encoder and the temperature $\tau$ in infoNCE loss is 0.1. As implemented in \cite{he2020momentum}, shuffling BN is employed to avoid data leaks. For the optimization of video encoder, we use the momentum SGD with an initial learning rate 0.01 which is annealed down to zero via a cosine decay. Following \cite{yao2020seco}, the network is trained for 400 epochs on Kinetics-400 and 200 epochs on UCF101 with the network initialization by MoCo on ImageNet.

\textbf{Linear Protocol.} We directly utilize the backbone pre-learnt by MCL as a feature extractor and examine the quality of frozen features on UCF101 and Kinetics-400 datasets. Specifically, we densely sample 20 clips from each video and resize each clip with the short edge of 256. The resized clip is then cropped to $224\times224$ via the standard center crop, which is fed into the feature extractor to output the clip-level feature, and the video-level feature is the average of all clip-level features.
Finally, we train a linear SVM (UCF101) or a fully-connected layer followed by softmax (Kinetics-400) on the training set, and evaluate the performance on the corresponding validation set. The top-1 classification accuracy is adopted as the metric.

\textbf{Downstream Task Evaluations.} We use the MCL pre-trained backbone as network initialization for two downstream tasks of action recognition and video retrieval in order to examine the transfer capability of the pre-learnt structure. For action recognition, we initialize S3D or R(2+1)D network with the backbone self-supervised trained by MCL, and then fine-tune the backbone plus fully-connected layers in standard supervised setting on UCF101 and HMDB51. For video retrieval, we follow the common practice~\cite{xu2019self}, and use the representations of videos from test set to query the $k$-nearest neighbours in training set. We directly capitalize on the MCL pre-trained backbone without further fine-tuning to extract video representations. We adopt the Recall at $k$ (R@$k$) as the performance metric, and a correct retrieval is counted when the top-$k$ nearest neighbours contain at least one video from the category of the~query~video.

\begin{table}[!tb]
       \centering
       \caption{\small Performance contribution of each design in MCL with different backbone networks (All models are learnt on UCF101 and evaluated under linear protocol). }
       \vspace{-0.1in}
       \small
       \setlength{\tabcolsep}{1.7mm}
       \begin{tabular}[t]{cc|ccc|cc|cc}
              \hline
              \multirow{2}{*}{TA} & \multirow{2}{*}{SA} & \multirow{2}{*}{$\mathcal{L}_{t}$} & \multirow{2}{*}{$\mathcal{L}_{s}$} & \multirow{2}{*}{$\mathcal{L}_{st}$} & \multicolumn{2}{c|}{ResNet-50} & \multicolumn{2}{c}{Inception V1} \\
              \cline{6-9}
              &                                &                               &                     &                                     &   R2D & R(2+1)D                      & I2D & S3D\\
              \hline
              &              &              &                 &               &  76.51  &  77.98           &    74.39 & 76.47 \\
              $\surd$   &              &               &                &                & 76.47  &  78.26          &    74.22  &  76.79    \\
              & $\surd$ &               &                &               & 77.32  &  79.54            &   75.02  &  77.95    \\
              $\surd$   & $\surd$ &               &                &               &  77.16  &  79.64           &    74.89  &      78.01   \\
              \hline
              $\surd$   & $\surd$ & $\surd$  &                &               &  -        &  80.26           &     -       &      78.62  \\
              $\surd$   & $\surd$ &               & $\surd$  &                & \textbf{77.84}  & 80.83 &\textbf{75.64} &     79.28      \\
              $\surd$   & $\surd$ &               &               & $\surd$   &    -       & 81.47            &    -        &    79.40      \\
              $\surd$   & $\surd$  &$\surd$  & $\surd$  & $\surd$   &    -       &\textbf{81.91} &     -       &    \textbf{79.88}   \\
              \hline
       \end{tabular}
       \label{main_ablation}
    \vspace{-0.20in}
\end{table}

\subsection{An Ablation Study of MCL}
We first examine how each design in MCL impacts the overall performance. The baseline of data augmentation randomly samples spatial or spatio-temporal patches from the whole video. Temporal Augmentation (TA) solely exploits the clip-level motion to choose the video clips with relatively large motion and randomly localizes the spatial patches in those clips. In contrast, Spatial Augmentation (SA) randomly selects video clips but employs S-motion to locate spatial patches. $\mathcal{L}_{s}$, $\mathcal{L}_{t}$, and $\mathcal{L}_{st}$ denotes the three items in Eq.(\ref{eq:mal}) and leverages the motion alignment loss from spatial, temporal and spatio-temporal standpoint, respectively. Please note that R2D and I2D are both 2D backbone networks, and the inputs to the two networks are downgraded from 3D tubelets to 2D patches on single frames. As a result, only the grounding on S-motion, i.e., $\mathcal{L}_{s}$, is applicable in this case. 

Table \ref{main_ablation} summarizes the top-1 accuracy on UCF101 dataset under linear protocol by considering different factors in MCL with various backbone networks. The results consistently indicate that capitalizing on motion to achieve augmentations in SA exhibits performance boost against augmentation by random sampling across 2D and 3D network backbones. Interestingly, TA leads to improvements on 3D network backbones but results in slight drop on 2D network backbones. We speculate that this may be the result of drastic appearance changes in the frames selected by TA and taking the patches from such single frames as inputs to 2D networks may adversely affect feature learning. Involving both TA and SA in MCL contributes a further increase of accuracy with 3D networks of R(2+1)D and~S3D. The comparisons also demonstrate the advantages of grounding motion information in feature learning. As aforementioned, only the alignment in spatial manner, i.e., $\mathcal{L}_{s}$, fits 2D backbone networks and boosts up the accuracy from 77.16\%/74.89\% to 77.84\%/75.64\% on R2D/I2D. Moreover, allowing motion grounding in terms of $\mathcal{L}_{t}$, $\mathcal{L}_{s}$, and $\mathcal{L}_{st}$ leads to performance gains in the range of 0.62\%, 1.19\% and 1.83\% on R(2+1)D, and 0.61\%, 1.27\% and 1.39\% on S3D. As expected, executing $\mathcal{L}_{st}$ outperforms the use of $\mathcal{L}_{s}$ or $\mathcal{L}_{t}$. By fusing the three losses, the accuracy of MCL finally reaches 81.91\% and 79.88\% with the backbone of R(2+1)D and S3D.

\begin{table}[!tb]
       \centering
       \caption{\small Comparisons of different ways on motion alignment (All models are learnt on UCF101 and evaluated under linear protocol).}
    \small
       \vspace{-0.1in}
       \begin{tabular}[t]{c|c}
              \hline
              Method  & Top-1 Acc. (\%)  \\
              \hline
              MCL w/o $\mathcal{L}_{\text{MAL}}$       & 79.64 \\
              MCL w/ $\mathcal{L}_{\text{MAL-v1}}$    & 80.57 \\
              MCL w/ $\mathcal{L}_{\text{MAL-v2}}$    & 80.99 \\
              MCL w/ $\mathcal{L}_{\text{MAL-v3}}$    & \textbf{81.91} \\
              \hline
       \end{tabular}
       \label{grad}
    \vspace{-0.20in}
\end{table}

\begin{table*}[!tb]
       \centering
       \caption{\small Performance comparisons of video representations pre-learnt by different approaches on UCF101 and Kinetics-400 under linear protocol. * represents the results based on the pre-trained models released by original publications on Github. The training time is estimated on one Nvidia P40 GPU. The \# of parameters and Flops are counted on an input image/clip with the resolution used in original~publications.}
       \vspace{-0.07in}
       \small
       \begin{tabular}[t]{l|l|c|c|c|c|c|c}
              \hline
              Method & Network & Pretrain Dataset  & \#Param. & FLOPs & Training Time & Eval  Dataset   & Top1 Acc.(\%) \\
              \hline
              PRP~\cite{yao2020video}& R(2+1)D & UCF101& 14.4M & 21.5G &  -  &UCF101  & 32.10$^{\ast}$ \\
              IIC~\cite{tao2020self}& R3D-10 & UCF101& 14.4M & 19.9G & - &UCF101  & 35.13$^{\ast}$ \\
              CCL~\cite{kong2020cycle}& R3D-18+1 & Kinetics-400& 12.1M & 16.4G& - &UCF101  & 52.10 \\
              MemDPC~\cite{han2020memory} & R-2D3D-34 & Kinetics-400 & 32.4M & 25.5G&  -  & UCF101  & 54.10 \\
              TCLR~\cite{dave2021tclr}& R3D-18 & UCF101 & 33.0M & 32.9G &  - &UCF101  & 67.70 \\
              CoCLR~\cite{han2020self} & S3D & UCF101 & 7.9M & 12.0G & 2.3 days&UCF101  & 70.20 \\
              ImageNet infla. & R(2+1)D & ImageNet &  27.3M & 20.4G & - &UCF101  & 75.13 \\
              Supervised ImageNet & R-50 & ImageNet & 23.5M  & 4.12G & - &UCF101  & 73.24 \\
              SeCo~\cite{yao2020seco} & R-50 & UCF101 & 23.5M &4.12G & 0.3 days& UCF101  & 76.51 \\
              SeCo~\cite{yao2020seco} & R(2+1)D & UCF101 & 27.3M & 20.4G & 1.4 days& UCF101  & 77.98 \\
              \hline
              MCL (Ours) & S3D & UCF101 & 7.9M & 18.4G & 2.1 days&UCF101  & \textbf{79.88} \\
              MCL (Ours) & R(2+1)D & UCF101 & 27.3M & 20.4G & 1.5 days &UCF101  & \textbf{81.91} \\
              \hline
              \hline
              VTHCL~\cite{yang2020video} & R3D-50 & Kinetics-400 & 31.7M & -& - &Kinetics-400  & 37.83 \\
              SimCLR infla.~\cite{qian2020spatiotemporal} & R3D-50 & Kinetics-400 & 31.7M  & 45.8G& - &Kinetics-400  & 46.80 \\
              VINCE~\cite{gordon2020watching} & R-50 & Kinetics-400 & 23.5M  & 4.12G & 17.6 days&Kinetics-400  & 49.10 \\
              ImageNet infla.~\cite{qian2020spatiotemporal} & R3D-50 & Kinetics-400 & 31.7M  & 45.8G& - &Kinetics-400  & 53.50 \\
              SeCo~\cite{yao2020seco} & R-50 & Kinetics-400 & 23.5M  & 4.12G & 13.0 days &Kinetics-400  & 61.91 \\
              SeCo~\cite{yao2020seco} & R(2+1)D & Kinetics-400 & 27.3M & 20.4G & 76.0 days &Kinetics-400  & 62.50 \\
              CVRL~\cite{qian2020spatiotemporal} & R3D-50 & Kinetics-400 & 31.7M & 45.8G & 322.6 days & Kinetics-400  & 66.10 \\
              \hline
              MCL (Ours) & R(2+1)D & Kinetics-400 & 27.3M & 20.4G & 76.5 days &Kinetics-400  &  \textbf{66.62} \\
              \hline
       \end{tabular}
       \label{linear_eval}
    \vspace{-0.15in}
\end{table*}

\subsection{Evaluations on Motion Alignment}
We then study the effect of three different ways as defined in Eq.(\ref{eq:mal-1}), Eq.(\ref{eq:mal-2}) and Eq.(\ref{eq:mal-3}) for motion alignment in MCL. Table \ref{grad} shows the performance comparisons across the three kinds of alignments. As indicated by the results, the use of motion alignment favors the representation learning. That empirically validates the grounding of motion information in MCL. Among the three ways of alignments, MCL w/ $\mathcal{L}_{\text{MAL-v3}}$ benefits from the explicit impact on gradient map and leads to a larger performance gain.

\subsection{Evaluations on Linear Protocol}
Next, we evaluate MCL under linear protocol to verify the representations learnt by MCL. Table \ref{linear_eval} details performance comparisons of different representation learning methods on UCF101 and Kinetics-400 datasets. Overall, our MCL leads to a performance boost against all the other baselines on UCF101. In particular, performing classification on the representations pre-learnt by MCL with the backbone of S3D and R(2+1)D achieves 79.88\% and 81.91\%, respectively.Compared to self-supervised method CoCLR with S3D backbone, MCL makes the absolute performance improvement by 9.68\% based on the same backbone.Moreover, MCL leads the top-1 accuracy by 3.93\% over the best competitor SeCo based on the same R(2+1)D backbone. The results empirically verify the idea of leveraging motion in MCL for self-supervised video representation learning. Similar to the observations on UCF101, pre-training MCL on Kinetics-400 dataset outperforms the baselines. 
MCL with R(2+1)D backbone obtains 66.62\% top-1 accuracy and leads to 4.12\% performance gain over SeCo with the same backbone. 
Compared to CVRL which requires 4$\times$ the training time and 2$\times$ FLOPs of MCL, MCL also leads to an accuracy boost of 0.52\%.

\begin{table}[htb]
       \centering
       \small
       \setlength{\tabcolsep}{0.5mm}
       \caption{Performance comparisons on UCF101 and HMDB51 for downstream task of action recognition.}
       \vspace{-0.08in}
       \begin{tabular}[t]{l|l|c|cc}
              \hline
              Method & Network & Pretrain Dataset  & UCF101 & HMDB51 \\
              \hline
              OPN~\cite{lee2017unsupervised} & VGG & UCF101 &   59.60 & 23.80\\
              VCOP~\cite{xu2019self} & R(2+1)D & UCF101 &   72.40 & 30.90\\
              CoCLR~\cite{han2020self} & S3D & UCF101 & 81.40 & 52.10\\
              BE~\cite{wang2020removing} & R3D-34 & UCF101 &    83.40 & 53.70\\
              SeCo~\cite{yao2020seco} & R-50 & UCF101 & 83.39  & 50.19 \\
              SeCo~\cite{yao2020seco} & R(2+1)D & UCF101 &89.82  & 56.40  \\
              \hline
              MCL (Ours) & S3D & UCF101 & \textbf{90.58}  & \textbf{63.52}  \\
              MCL (Ours) & R(2+1)D & UCF101 &\textbf{90.40}  & \textbf{61.30}   \\
              \hline
              \hline
              3D-RotNet~\cite{jing2018self} & R3D-18 & Kinetics-400 &  62.90 & 23.80\\
              ST-Puzzle~\cite{kim2019self} & R3D-18 & Kinetics-400 &  63.90 & 33.70\\
              DPC~\cite{han2019video}  & R-2D3D-34 & Kinetics-400 & 75.70 & 35.70\\
              MemDPC~\cite{han2020memory} & R-2D3D-34 & Kinetics-400 & 78.10 & 41.20\\
              SpeedNet~\cite{benaim2020speednet} & S3D & Kinetics-400 &  81.10 & 48.80\\
              CoCLR~\cite{han2020self} & S3D & Kinetics-400 &87.90 & 54.60\\
              BE~\cite{wang2020removing} & R3D-34 & Kinetics-400 &  87.10 & 56.20\\
              SeCo~\cite{yao2020seco} & R-50 & Kinetics-400 &  88.26 & 55.55\\
              CVRL~\cite{qian2020spatiotemporal} & R3D-50 & Kinetics-400 &  92.20 & 66.70 \\
              CBT~\cite{sun2019learning} & S3D & K600+ & 79.50 & 44.60\\
              DynamoNet~\cite{diba2019dynamonet} & STCNet & YouTube8M-1 &  88.10 & 59.90\\
              \hline
              MCL (Ours) & R(2+1)D & Kinetics-400 &\textbf{93.41}  & \textbf{69.08}   \\
              \hline
       \end{tabular}
       \label{downstream:classification}
    \vspace{-0.25in}
\end{table}

\begin{table*}[htb]
       \centering
       \setlength{\tabcolsep}{0.8mm}
       \caption{Performance comparisons on UCF101 and HMDB51 for downstream task of video retrieval.}
       \small
       \begin{tabular}[t]{l|l|c|cccc|cccc}
              \hline
              \multirow{2}{*}{Method} & \multirow{2}{*}{Network} & \multirow{2}{*}{Pretrain Dataset}&\multicolumn{4}{c}{UCF101} &\multicolumn{4}{c}{HMDB51}\\
              \cline{4-11}
              &              &             & R @ 1 & R @ 5 & R @ 10 & R @ 20 & R @ 1 & R @ 5 & R @ 10 & R @ 20 \\
              \hline
              OPN~\cite{lee2017unsupervised}                & VGG      & UCF101 &   19.9    &   28.7  &   34.0  &  40.6   &    -     &     -     &    -       &     -      \\
              VCOP~\cite{xu2019self}             & R3D-18 & UCF101 &   14.1    &   30.3  &   40.4  &   51.1    &    7.6     &     22.9     &    34.4       &     48.8      \\
              VCP~\cite{luo2020video}                 & R3D-18 & UCF101 &   18.6   &   33.6  &   42.5  &  53.5    &    7.6     &     24.4     &    36.3       &   53.6      \\
              MemDPC~\cite{han2020memory}          & R-2D3D-18 & UCF101 &   20.2   &   40.4  &   52.4  &  64.7    &    7.7     &     25.7     &    40.6      &    57.7       \\
              SpeedNet~\cite{benaim2020speednet}         & S3D-G  & Kinetics-400    &   13.0   &   28.1  &   37.5  &  49.5    &    -    &     -     &    -      &    -       \\
              PRP~\cite{yao2020video}                  & R(2+1)D & UCF101 &   20.3   &   34.0  &   41.9  &  51.7    &    8.2     &     25.3     &    36.2      &    51.0       \\
              BE~\cite{wang2020removing}    &R3D-34 &UCF101  &    -    &     -     &    -      &    -    &11.9 &31.3 &44.5 &60.5 \\
              CoCLR-RGB~\cite{han2020self}      & S3D     & UCF101  &   53.3   &   69.4  &  76.6  &  82.0   &    23.2     &    43.2     &    53.5      &    65.5       \\
              CoCLR-2Stream~\cite{han2020self} & S3D    & UCF101  &    55.9  &  70.8   &   76.9 &   82.5 &  26.1 &  45.8 & 57.9 & 69.7 \\
              \hline
              MCL (Ours)                & S3D     & UCF101  &  \textbf{67.0}  &   \textbf{80.8}  &  \textbf{86.3}  &  \textbf{90.8}   &    \textbf{26.7}    &    \textbf{52.5}    &   \textbf{67.0}       &    \textbf{79.3}      \\
              MCL (Ours)                & R(2+1)D & UCF101  &\textbf{68.6}&\textbf{82.2}&\textbf{87.2}&\textbf{92.0}&\textbf{29.0}&\textbf{55.5}&\textbf{68.9} &\textbf{80.4}\\
              \hline
       \end{tabular}
       \label{downstream:retrieval}
    \vspace{-0.15in}
\end{table*}

\subsection{Evaluations on Downstream Tasks}
Another common protocol in self-supervised learning is to take the network pre-training as network initialization and fine-tune all layers on downstream tasks. Table \ref{downstream:classification} shows the comparisons of pre-training the networks by different models and then supervised fine-tuning on UCF101 and HMDB51 for action recognition, which are the most widely-adopted evaluations in the literature. The results across pre-training on UCF101 and Kinetics-400 datasets constantly indicate that our MCL exhibits better performances than all the baselines. Fine-tuning the networks pre-trained on UCF101 by MCL achieves 90.58\% and 63.52\% on UCF101 and HMDB51, respectively, leading to apparent improvements over VCOP, CoCLR and SeCo. Compared to SeCo which reports the best known results, MCL with R(2+1)D backbone boosts up the accuracy from 89.82\%/56.40\% to 90.40\%/61.30\% on UCF101 and HMDB51 datasets. Pre-training the networks on the larger Kinetics-400 dataset by MCL further improves the accuracy on UCF101 and HMDB51 to 93.41\% and 69.08\%, and leads the accuracy by 1.21\% and 2.38\% against CVRL. Notably, MCL is also superior to DynamoNet pre-trained on a subset of YouTube-8M with 2$\times$ size of Kinetics-400, which is~impressive.

Table \ref{downstream:retrieval} summarizes the comparisons on UCF101 and HMDB51 for video retrieval task, which is to examine the semantics of $k$ training videos nearest to the query video from test set in the representation space. As indicated by the results across different depths of Recall, MCL yields higher scores than other methods on the two datasets. Taking S3D as the backbone, MCL with only RGB inputs still leads the Recall@1 score by 11.1\% and 0.6\% over CoCLR with the two-stream inputs of RGB and optical flow modalities on UCF101 and HMDB51. The use of more powerful R(2+1)D backbone in MCL further contributes a Recall@1 increase of 1.6\% and 2.3\%. The results successfully demonstrate the transferability of the pre-trained structure by MCL to different downstream tasks.
\begin{figure}[!tb]
       \centering
       \includegraphics[width=0.43\textwidth]{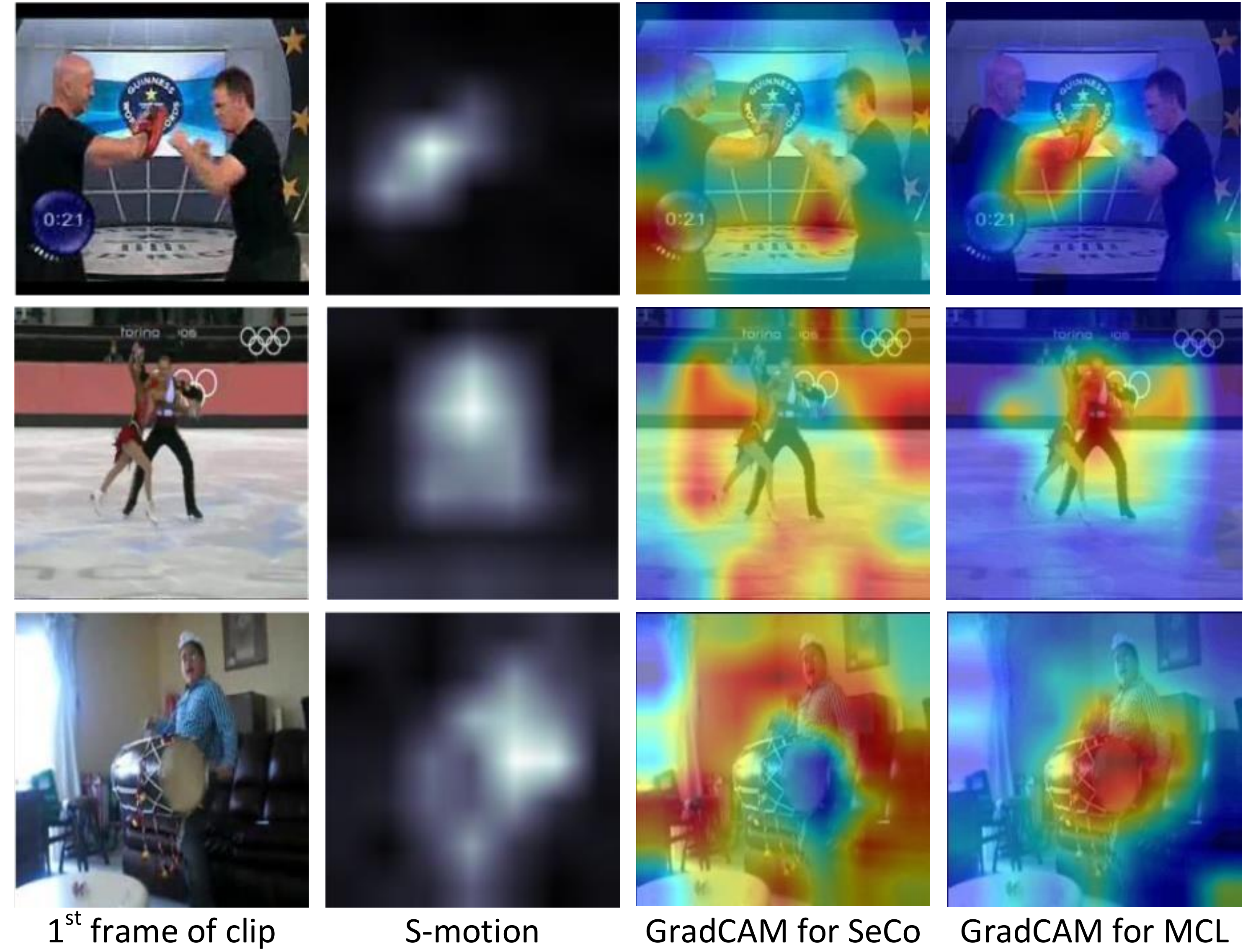}
       \caption{\small From left to right: the first frame from video clip, S-motion, Grad-CAM for SeCo, Grad-CAM for MCL. }
       \label{visualization}
    \vspace{-0.15in}
\end{figure}

\subsection{Visualizing Self-supervised Representation}
In order to explain the representations learnt by MCL, we generate the saliency map through GradCAM~\cite{selvaraju2017grad} to indicate the importance of each pixel in presenting the discrimination of the video clip. Figure \ref{visualization} visualizes the start frame of a video clip, S-motion of the clip, the saliency map produced on the representations learnt by SeCo and our MCL with R(2+1)D backbone. Note that we compute a 3D saliency map on the video clip and perform temporal pooling to depict the map here. MCL benefits from the leverage of motion and the saliency map on the representations learnt by MCL aligns S-motion more nicely than that by SeCo. More importantly, the regions of high importance effectively provide the visual evidence for describing the video clip and thus the representations learnt by MCL are potentially more robust. That again proves the utilization of motion in MCL for self-supervised representation learning.
\vspace{-0.15in}
\section{Conclusion}
\vspace{-0.05in}
We have presented a Motion-focused Contrastive Learning (MCL) method, which explores the motion information for improving self-supervised video representation learning. Particularly, we study the problem via leveraging motion to achieve data augmentations and enhance feature learning in contrastive learning framework. To materialize our idea, we extract the motion map of each frame and form a sequence of motion maps in a video clip as the spatio-temporal motion map of the clip. The output value of 3D average pooling on the spatio-temporal motion map is taken as the measure of clip-level motion, which acts as an indicator to select the clip candidates for temporal augmentation. Then, we perform temporal pooling on spatio-temporal motion map of each clip candidate to estimate the motion of every spatial position along the time and localize the spatial patches temporally consistent across frames, as tubelets. MCL employs such tubelets as data augmentations for contrastive learning and further aligns gradients of the convolutional layers to motion maps of the tubelets from spatial, temporal and spatio-temporal aspects. Extensive experiments on UCF101 and Kinetics-400 datasets validate our MCL. More remarkably, self-supervised pre-training MCL is superior to fully-supervised ImageNet pre-training.

\textbf{Acknowledgments.} This work was supported in part by the Natural Science Foundation of China under Grants 62036005 and 62021001, and in part by the Fundamental Research Funds for the Central Universities under contract WK3490000005.

{\small
\bibliographystyle{ieee_fullname}
\bibliography{egbib}
}

\end{document}